%% file: main.tex
\documentclass[runningheads]{llncs}
\usepackage[T1]{fontenc}
\usepackage{graphicx}
\usepackage{booktabs}
\usepackage[misc]{ifsym}
\newcommand{\corr}{(\Letter)}

\usepackage{hyperref}
\usepackage{multirow}
\usepackage{graphicx}
\usepackage[boxed,ruled,lined,linesnumbered]{algorithm2e}
\usepackage{wrapfig}
\usepackage[acronym]{glossaries}
\glsdisablehyper

\newacronym{abb:ml}{ML}{Machine Learning}
\newacronym{abb:fl}{FL}{Federated Learning}
\newacronym{abb:gnn}{GNN}{graph Neural Network}
\newacronym{abb:mf}{MF}{Matrix Factorization}
\newacronym{abb:ncf}{NCF}{Neural Collaborative Filtering}
\newacronym{abb:fedsgd}{FedSGD}{federated learning via stochastic gradient descent}
\newacronym{abb:fedavg}{FedAvg}{federated averaging}

\newcommand{\subfour}[1]{\vspace*{3mm}{\noindent\bf #1}}

\begin{document}
\title{Far From Sight, Far From Mind: Inverse Distance Weighting for Graph Federated Recommendation}

\titlerunning{Inverse Distance Weighting for Graph Federated Recommendation}

\author{Aymen Rayane Khouas\inst{1}\orcidID{0009-0008-8596-4154} \corr \and
 Mohamed Reda Bouadjenek\inst{1}\orcidID{0000-0003-1807-430X} \and
 Hakim Hacid\inst{2}\orcidID{0000-0003-2265-9343} \and
Sunil Aryal\inst{1}\orcidID{0000-0002-6639-6824}}

\authorrunning{A.R. Khouas et al.}

\institute{
  Deakin University, 75 Pigdons Road, Waurn Ponds, VIC, AU \email{\{a.khouas,reda.bouadjenek,sunil.aryal\}@deakin.edu.au}
  \and
  TII, UAE \email{hakim.hacid@tii.ae}
}

\maketitle

\begin{abstract}
  Graph federated recommendation systems offer a privacy-preserving alternative to traditional centralized recommendation architectures, which often raise concerns about data security. While federated learning enables personalized recommendations without exposing raw user data, existing aggregation methods overlook the unique properties of user embeddings in this setting.
  Indeed, traditional aggregation methods fail to account for their complexity and the critical role of user similarity in recommendation effectiveness. 
  Moreover, evolving user interactions require adaptive aggregation while preserving the influence of high-relevance anchor users—the primary users before expansion in graph-based frameworks.
  To address these limitations, we introduce \textit{Dist-FedAvg},  a novel distance-based aggregation method designed to enhance personalization and aggregation efficiency in graph federated learning.  
  Our method assigns higher aggregation weights to users with similar embeddings, while ensuring that anchor users retain significant influence in local updates. 
  Empirical evaluations on multiple datasets demonstrate that Dist-FedAvg consistently outperforms baseline aggregation techniques, improving recommendation accuracy while maintaining seamless integration into existing federated learning frameworks.
  
\end{abstract}

\input{sources/introduction}
\input{sources/related_works}

\input{sources/proposed_method}

\input{sources/experiments}

\input{sources/conclusion}

\begin{credits}
\subsubsection{\ackname} This research is supported by the Technology Innovation Institute, UAE under the research contract number TII/DSRC/2022/3143.
\end{credits}

\bibliographystyle{splncs04}
\bibliography{biblio}

\appendix

\end{document}

%% file: sources/introduction.tex
\section{Introduction}\label{sec:introduction}

Recommendation Systems (RS) have become an essential component of modern online services, helping users navigate vast amounts of information and discover relevant content~\cite{luRecommenderSystemApplication2015}. Traditional recommendation systems rely on centralized architectures, where user data is collected and processed in a cloud-based infrastructure. However, these systems face growing challenges associated to data security, user privacy and regulatory compliance~\cite{yinDeviceRecommenderSystems2024}.
In this context, Federated Recommendation Systems (FRS) have emerged as a promising alternative~\cite{sunSurveyFederatedRecommendation2024}, based on the premise that each client represents a single user and their unique interactions.
By enabling the training of recommendation models directly on user devices without sharing or transmitting local interactions, it enhances privacy and ensures compliance with data protection regulations. Furthermore, FRS leverages distributed computation to minimize reliance on centralized infrastructure, offering improved scalability and resilience against single points of failure.

In recent years, several methods have been proposed for FRS, ranging from federated \Gls{abb:mf} techniques~\cite{chaiSecureFederatedMatrix2021} and federated \Gls{abb:ncf}~\cite{perifanisFederatedNeuralCollaborative2022} to federated \Glspl{abb:gnn}-based methods~\cite{wuFederatedGraphNeural2022,zhangGPFedRecGraphGuidedPersonalization2024}. 
The latter is particularly promising given the significant impact of \Glspl{abb:gnn} in centralized recommendation settings~\cite{wuGraphNeuralNetworks2022a}, especially because of their ability to capture high-order relations between users and items. 
At the heart of federated \Glspl{abb:gnn} are aggregation methods, which combine locally updated models by aggregating their parameters to form a global model~\cite{mcmahanCommunicationEfficientLearningDeep2017}.
Nonetheless, several open challenges remain in developing advanced aggregation methods that can adapt to user similarity metrics, manage dynamic user-item interactions, and incorporate real-time updates seamlessly.

Multiple aggregation techniques have been proposed over the years~\cite{qiModelAggregationTechniques2024}, but we argue that they often neglect the unique characteristics of recommendation parameters, especially in graph-based settings.
Indeed, while item and model parameters are relatively uniform and can easily be aggregated using conventional methods, user parameters (or embeddings) are more complex and specialized aggregation approaches can yield significant benefits.
Additionally, user similarity plays a crucial role in recommendation effectiveness, making it essential to prioritize contributions from similar users~\cite{shiExploitingUserSimilarity2009}. 
Finally, due to the dynamic nature of user interactions, aggregation methods must adapt to evolving embeddings while preserving the influence of anchor users—the primary users prior to expansion in graph-based frameworks—who hold higher relevance in local training.
Therefore, user similarity must be considered during aggregation to ensure that meaningful relationships are accurately captured in the global model.

This paper addresses these limitations by introducing \textit{Dist-FedAvg}, a novel distance-based aggregation algorithm for user parameters in graph federated learning designed for seamless integration into existing frameworks. 
Dist-FedAvg assigns weights to user embeddings based on user similarity, prioritizing those with closer embeddings. 
It combines local updates from different clients by accounting for both user parameter similarity and the contributions of anchor users, enabling more meaningful and personalized aggregation in FRS.
We evaluate the performance of Dist-FedAvg on five different datasets, Movie Lens 100k, Movie Lens 1M, LastFM 2k, Amazon Digital Music and FilmTrust.
Experiments on these datasets show that our proposed Dist-FedAvg function improves performance compared to several baselines and state-of-the-art averaging methods for federated recommendation algorithms.
Our contributions are as follows:
\begin{itemize}
    \item We propose a novel distance-based aggregation function for user parameters in a graph federated problem, which can be seamlessly integrated with existing frameworks. 
    \item We introduce an anchor-user-aware aggregation strategy, which preserves the influence of primary users in local training while adapting to evolving user embeddings.
    \item We provide a comprehensive evaluation of Dist-FedAvg on five benchmark datasets, as well as an ablation study of our method.
    Experimental results demonstrate that our approach outperforms several baseline and state-of-the-art aggregation methods.
\end{itemize}

%% file: sources/related_works.tex
\section{Related Work}
\label{sec:related-work}
Federated recommendation systems have gained significant attention as a privacy-preserving alternative to traditional centralized approaches. 
In this section, we review key developments in this area, covering fundamental recommendation techniques, the evolution of \Gls{abb:fl} for recommendation, and recent advancements in federated graph-based models. 
We also examine aggregation methods in \Gls{abb:fl}, which play a crucial role in model performance and adaptation.

\subfour{Recommendation systems:} 
Recommender systems have proven to be an important set of techniques for filtering information and recommending relevant items based on users’ interaction histories. Collaborative filtering, which recommends items based on the similarity between users, is a widely used technique.
While earlier collaborative recommendation approaches focused on clustering, decision trees, and association rule mining algorithms~\cite{isinkayeRecommendationSystemsPrinciples2015}, modern collaborative approaches are usually based on \Gls{abb:mf}~\cite{korenMatrixFactorizationTechniques2009} or deep learning methods such as \Gls{abb:ncf}~\cite{heNeuralCollaborativeFiltering2017}, Autoencoders~\cite{sedhainAutoRecAutoencodersMeet2015,liangVariationalAutoencodersCollaborative2018}, two-tower systems~\cite{huangLearningDeepStructured2013} and \Glspl{abb:gnn}-based recommendation~\cite{wuGraphNeuralNetworks2022a,wangNeuralGraphCollaborative2019,heLightGCNSimplifyingPowering2020}, which emerged as prominent techniques in the field, achieving state-of-the-art performance. 
\Glspl{abb:gnn}-based RS have shown great promise in the past few years, with methods such as NGCF~\cite{wangNeuralGraphCollaborative2019}, LightGCN~\cite{heLightGCNSimplifyingPowering2020}, LightGT~\cite{weiLightGTLightGraph2023}, and others. The success of \Glspl{abb:gnn} is partially due to their ability to capture high-order relations between users and items~\cite{wuGraphNeuralNetworks2022a}, in contrast to \Gls{abb:mf}-based methods that only learn from first-order user-item interactions~\cite{flanaganFederatedMultiviewMatrix2021}. 

\subfour{Federated recommendation:} 
Unlike content-based filtering, which recommends items based on item features, collaborative filtering learns from similar users. 
Since each client in a federated setting holds the interaction data of a single user, it is necessary to design specific schemes for collaborative training. 
One of the earliest works in federated collaborative filtering is~\cite{ammad-ud-dinFederatedCollaborativeFiltering2019}, which introduced a method with federated updates based on a stochastic gradient approach. FedMF~\cite{chaiSecureFederatedMatrix2021} integrates \Gls{abb:mf} with \Gls{abb:fl}, proposing a user-level distributed \Gls{abb:mf} framework where each user only uploads gradient information. 
Other federated \Gls{abb:mf}-based approaches for collaborative recommendation include~\cite{yingSharedMFPrivacypreserving2020,duFederatedMatrixFactorization2021,yangPracticalSecureFederated2022}. 
Finally, FedNCF~\cite{perifanisFederatedNeuralCollaborative2022} proposes a federated variant of \Gls{abb:ncf}. 
The model combines a generalized \Gls{abb:mf} component with a multilayer perceptron (MLP) to train the recommendation parameters. 
While the parameters include the MLP’s weights as well as user and item profiles, only the MLP’s parameters and item profiles are sent to the server for aggregation. 
Another \Gls{abb:ncf}-inspired method, FNCF~\cite{aliwaqarCommunicationEfficientFederatedNeural2023}, employs multi-armed bandits to select a smaller set of payloads in each iteration of the federated training process, addressing model complexity challenges.

\subfour{Federated graph recommendation:} 
FedPerGNN~\cite{wuFederatedGraphNeural2022} is the first method proposed for federated graph recommendation. 
FedPerGNN introduces a graph expansion algorithm to expand local users beyond the client and enable training on higher-order user-item relations. 
The expansion algorithm relies on a trusted third-party server that matches users based on encrypted shared interactions. 
FedPerGNN also proposes a pseudo-item sampling along with a differential privacy strategy to ensure user privacy. 
Various methods have been built upon FedPerGNN, such as FeSoG~\cite{liuFederatedSocialRecommendation2022} for social recommendation and ShuffledFedGNN~\cite{liuPrivacyPreservingRecommendationBased2024} that proposes a shuffling mechanism instead of pseudo-item sampling used in FedPerGNN.
Another federated graph recommendation approach is GPFedRec~\cite{zhangGPFedRecGraphGuidedPersonalization2024}, which introduces a method for constructing a user-relation graph without accessing users' interactions. 
GPFedRec also proposes a graph-guided aggregation mechanism for learning user-specific item embeddings. 
On the other hand, SemiDFEGL~\cite{quSemidecentralizedFederatedEgo2023} proposes an alternative to graph expansion by clustering users using ego graph embeddings trained locally by each client and training the message passing algorithm in a decentralized way. In other words, instead of being stored on a single device, the graph of user-item interactions is distributed across a cluster of devices. CDCGNNFed~\cite{quPersonalizedPrivacyUserGoverned2024} proposes a cloud-device collaborative framework in which users have the option to choose what data to send to the cloud and what data to keep private. The method combines the local data with user-centric ego graphs and the data sent to the server with high-order graphs. CdFed~\cite{zhangClusterdrivenGNNbasedFederated2023} proposes an adaptive clustering-based federated graph recommendation. Finally, FedHGNN~\cite{yanFederatedHeterogeneousGraph2024} proposes an approach based on heterogeneous information networks.

\subfour{Aggregation methods in federated learning:} 
One of the key steps in \Gls{abb:fl} is model aggregation, where local models from multiple clients are combined into a single global model in each training round~\cite{qiModelAggregationTechniques2024}. 
The most widely used aggregation method is FedAvg~\cite{mcmahanCommunicationEfficientLearningDeep2017}, defined as:
\vspace{-0.3cm}

\begin{equation}\label{eq:agg:fedavg}
    \mathsf{w}_{t+1} = \sum_{k=1}^{\left|\mathcal{C}^{(r)}\right|} \frac{n_{k}}{n} \mathsf{w}^{k}_{t+1},
\end{equation}
\vspace{-0.3cm}

with $\mathcal{C}^{(r)}$ representing the set of selected clients, $n_{k}$ the number of instances for client $k$, $n$ the total number of data instances, $\mathsf{w}^{k}_{t+1}$ the model weights of client $k$ at iteration $t+1$, and $\mathsf{w}_{t+1}$ the global model weights.
Multiple aggregation methods have been proposed to account for various factors, such as the quality of local models, the robustness of the aggregation against Byzantine attacks, the number of selected clients, and the heterogeneity of local data~\cite{qiModelAggregationTechniques2024}.
Custom aggregation methods have also been proposed for federated recommendation, such as FedFast~\cite{muhammadFedFastGoingAverage2020}, which introduces an active sampling and aggregation method to accelerate the convergence of local models. Additionally, GPFedRec~\cite{zhangGPFedRecGraphGuidedPersonalization2024} proposes an item aggregation scheme for user-specific item embeddings in the graph federated recommendation setting.

%% file: sources/proposed_method.tex
\begin{table}[t!]
\centering
    \caption{Key Notations}
    \label{tab:agg:notations}
    \resizebox{\columnwidth}{!}{%
    \begin{tabular}{l|l}
    \hline
    \textbf{Notation} & \textbf{Description} \\ \hline
    $\mathcal{U}$, $\mathcal{V}$, $\mathcal{C}$ & Respectively the set of users, items, and clients \\
    $u_i$, $v_j$, $c_i$ & Respectively user $i$, item $j$, and  client $i$ \\
    $u_{i,j}$ & Respectively user $j$ in client $i$\\
    $n$, $m$ & Respectively number of users/clients and items \\
    $\mathcal{C}^{(r)}$ & The set of selected clients for local training in round $r$ \\
    $\mathcal{G}$ & Graph of User-Item interactions \\
    $\mathcal{G}_i$ & Local Subgraph of User-Item interactions for client $i$ \\
    $\mathcal{U}_i$ & Set of expanded users into $c_i$\\
    $\mathbf{E}_u$, $\mathbf{E}_v$ & Respectively the global user and item embeddings matrices \\
    $\mathbf{e}_{u_i}$, $\mathbf{e}_{v_j}$ & Respectively global embeddings for user $u_i$ and item  $v_j$\\
    $\mathbf{E}_{u}^{(r)}$ / $\mathbf{E}_{v}^{(r)}$ & Respectively, the updated user and item matrix embeddings at round $r$ \\
    $\boldsymbol{\mathsf{E}}_{u}^{(r)}$ / $\boldsymbol{\mathsf{E}}_{v}^{(r)}$ & Respectively, 3D tensor of user/item embeddings collected from local clients \\ & before aggregation at round $r$\\
    $\mathbf{D}$ & The user-to-user distance matrix, where each entry $\mathbf{D}_{ij}$ represents the distance \\ & between user $i$ and user $j$ \\
    $\mathbf{W}$ & The matrix of averaging weights \\
    \hline
    \end{tabular}%
    }
\end{table}

\section{Preliminaries and Problem Formulation}
\label{sec:notation-problem-method}

\subsection{Graph Representation and notations}
Let $\mathcal{U} = \{u_1, \dots, u_n\}$ and $\mathcal{V} = \{v_1, \dots, v_m\}$ denote the sets of $n$ users and $m$ items, respectively.
We define the set of clients as $\mathcal{C} = \{c_1, \dots, c_n\}$, where each user in our cross-device FL setup corresponds to a unique client, so the number of clients is $n$.
Furthermore, let $\mathcal{G} = (\mathcal{U}, \mathcal{V}, \mathcal{E})$ represent the global graph of user-item interactions, where $\mathcal{E} \subseteq \mathcal{U} \times \mathcal{V}$ is the set of edges representing interactions (e.g., clicks, purchases, ratings) between users and items. 
Similarly, let $\mathcal{G}_i$ represents the local user-item graph at client $c_i$, which includes only the interactions between users and items within that specific client.
Initially, $\mathcal{G}_i$ represents the first-order user-item interactions of user $u_i$. 
Then, it is expanded to include additional users, denoted as $\mathcal{U}_i = \{u_{i,1}, \dots, u_{i,k}\}$, with $k$ representing the number of users in client $c_i$ after expansion. 
At round $r$, the user embeddings matrix is represented as $\mathbf{E}_{u}^{(r)}$ and the item embeddings matrix as $\mathbf{E}_{v}^{(r)}$. 
The vector embedding for user $u_i$ at round $r$ is denoted as $\mathbf{e}_{u_i}^{(r)}$.
Additionally, the embeddings of user $j$ expanded into client $i$ at round $r$ are denoted as $\mathbf{e}_{u_{i, j}}^{(r)}$. 
Finally, let $\boldsymbol{\mathsf{E}}_u^{(r)}$ denote the set of updated user embeddings received from all selected clients at round $r$. 
The selected clients are denoted as $\mathcal{C}^{(r)}$, which represent a subset of $\mathcal{C}$ selected for local training at a given round $r$. 
A summary of this paper's symbols is given in Table~\ref{tab:agg:notations}.

\subsection{Anchor user} 
The initial user $u_i$ in $c_i$ before the expansion, denoted $u_{i,i}$, is referred to as the \textbf{anchor}, \textbf{anchor user}, or \textbf{anchor client}.
It represents the central user around which the expanded graph is constructed. 
Each client $c_i$  has a single anchor user which serves as the central node for expansion. 
Furthermore, every expanded user $u_{i,j}$ originates from its corresponding anchor user, where $i=j$. 
Due to the anchor user's centrality, its embedding $\mathbf{e}_{u_{i,i}}$ has a greater contribution to local training and holds more relevance than any other embedding $\mathbf{e}_{u_{i,j}}$, where $u_i$ is not an anchor (i.e., $i \neq j$).
Because of that, anchor users have a great impact on the proposed aggregation strategy.

\subsection{Problem definition:}

In FRS, the effective aggregation of user-specific parameters is a significant challenge, particularly when leveraging user similarity. Our goal is to design an aggregation method that computes global user embeddings by weighting updated embeddings based on the similarity between users.
At the end of each round $r$, the central aggregation server receives a set of updated user/item embeddings $\boldsymbol{\mathsf{E}}_u^{(r)}$, from a selected subset of clients $\mathcal{C}^{(r)}$. We seek to derive the global user embeddings for the current round $\mathbf{E}_{u}^{(r)}$ from $\boldsymbol{\mathsf{E}}_u^{(r)}$, by incorporating information from the previous round’s global embeddings $\mathbf{E}_{u}^{(r-1)}$. This is achieved by weighting the updated embeddings according to their distance to the previous global embeddings. The aim of our method is formalized in the following equation: 
\begin{equation}\label{eq:agg:general_aggregation}
     \mathbf{E}_{u}^{(r)} = \operatorname{Agg}{\left(\boldsymbol{\mathsf{E}}_u^{(r)}, \mathbf{E}_{u}^{(r-1)}\right)}.
\end{equation}

\begin{algorithm}[t]
    \SetAlgoLined
    \KwIn{$\mathcal{U}$; $\mathcal{V}$; $\lambda$ : Local Training Hyperparameters; $\theta$ : Aggregation Hyperparameters}
    \KwOut{$\mathbf{E}_u$ and $\mathbf{E}_v$}
    
    \BlankLine
    Initialize $\mathbf{E}_u^{(0)}$ and $\mathbf{E}_v^{(0)}$\;
    Execute a user/graph expansion and create subgraphs $\mathcal{G}_i$\;
    \ForEach{round r=1, 2, ..., NumberOfRounds}{
        Select a set of clients $\mathcal{C}^{(r)}$ from $\mathcal{C}$\;
        $\boldsymbol{\mathsf{E}}_v^{(r)}$, $\boldsymbol{\mathsf{E}}_u^{(r)}$ $\leftarrow$ $\emptyset$\;
        \ForEach{client $ c_{j} \in \mathcal{C}^{(r)}$}{
            $\mathbf{E}^{(r)\prime}_{u_{j}}$, $\mathbf{E}^{(r)\prime}_{v_{j}}$ $\leftarrow$ $\operatorname{ClientUpdate}\left(\mathbf{E}_{u}^{(r-1)}, \mathbf{E}_{v}^{(r-1)}, \lambda\right)$\;
            $\boldsymbol{\mathsf{E}}_v^{(r)}$ $\leftarrow$ $\boldsymbol{\mathsf{E}}_v^{(r)}$ $\cup$ $\mathbf{E}^{(r)\prime}_{v_{j}}$\;
            $\boldsymbol{\mathsf{E}}_u^{(r)}$ $\leftarrow$ $\boldsymbol{\mathsf{E}}_u^{(r)}$ $\cup$ $\mathbf{E}^{(r)\prime}_{u_{j}}$\;
        }
        $\mathbf{E}_{u}^{(r)} \leftarrow \operatorname{AggregateUsersEmb}{\left(\boldsymbol{\mathsf{E}}_u^{(r)}, \mathbf{E}_{u}^{(r-1)}, \theta\right)}$\;
        $\mathbf{E}_{v}^{(r)} \leftarrow \operatorname{AggregateItemsEmb}{\left(\boldsymbol{\mathsf{E}}_v^{(r)}, \theta\right)}$\;
    }
    \caption{Optimization of Federated Graph Recommendation on Central Server}
    \label{algo:agg:central_server}
\end{algorithm}

We operate within a simplified framework inspired by FedPerGNN~\cite{wuFederatedGraphNeural2022}. In this framework, a trusted third-party server is responsible for user expansion by matching users based on encrypted shared interactions.
We ignore privacy preserving components such as pseudo-item sampling and differential privacy. While they are crucial for preserving user privacy, our goal is mainly to investigate the aggregation of user parameters\footnote{Since none of the methods rely on user-item interactions to aggregate user parameters, employing such privacy-preserving techniques does not affect their feasibility. However, we note that incorporating these techniques in practice may introduce noise, which could impact the quality of the embeddings.}.
Furthermore, to simplify the framework, we also operate under the assumption that the clients directly transmit the user embeddings $\mathbf{E}_u$ to the aggregation server. Although directly transmitting user parameters might raise privacy concerns, our method can be adapted for aggregating gradients while maintaining user embeddings' privacy by locally computing the distances and transmitting only the computed weights to the aggregation server.
A detailed description of the overall framework within which we operate is provided in Algorithm~\ref{algo:agg:central_server}.

\section{Proposed Aggregation Method}
\label{sec:proposed-method}

\begin{figure*}[t]
    \centering
    \includegraphics[width=\textwidth]{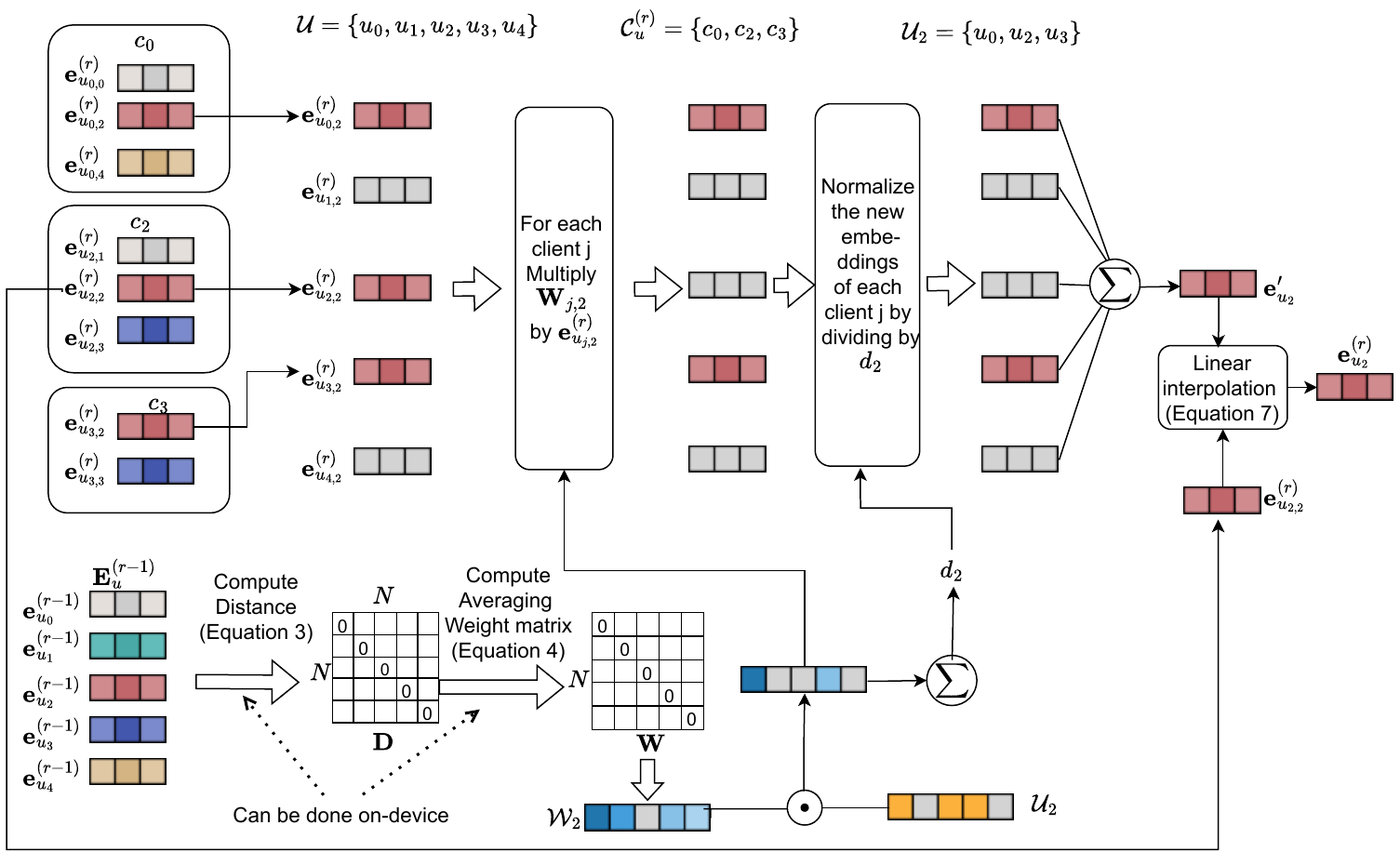}
    \caption{An illustration of the aggregation of embeddings for $u_2$ using Dist-FedAvg.}
    \label{fig:agg:dist_fedavg} 
\end{figure*}

In this section, we present Dist-FedAvg, an aggregation method that relies on the  normalized inverse distance between users and uses a linear interpolation with a dynamic $\alpha$ as its weighting scheme. We summarize the method for Dist-FedAvg in Algorithm~\ref{algo:agg:dist_fedavg} and provide a visual representation in Figure~\ref{fig:agg:dist_fedavg}.

\begin{algorithm}[t]
    \SetAlgoLined
    \KwIn{
        $\mathcal{U}$: set of users; $\mathbf{E}_{u}^{(r-1)}$: user embeddings from the previous round; $\boldsymbol{\mathsf{E}}_u^{(r)}$: updated user embeddings from clients; $\mathcal{C}^{(r)}$ : Selected clients for local training; $r$ : current round; $\alpha$: interpolation parameter (which can also be computed using Equation~\ref{eq:agg:inv_dist_compute_alpha} or~~\ref{eq:agg:inv_dist_compute_geo_alpha});
    }
    \KwOut{
        $\mathbf{E}_{u}^{(r)}$
    }
    \BlankLine

    If applying decay, compute the value of $\alpha$ in this round by using $\alpha_{0}$, $\alpha_{T}$, $\gamma$, $k$ and $r$ as presented in Equation~\ref{eq:agg:inv_dist_compute_alpha} or Equation~\ref{eq:agg:inv_dist_compute_geo_alpha}\;
    \ForEach{user $u_i$ $\in$ $\mathcal{U}$}{
        \ForEach{user $u_j$ $\in$ $\mathcal{U}$}{
            Compute distance between $\mathbf{e}^{(r-1)}_{u_i}$ and $\mathbf{e}^{(r-1)}_{u_j}$ using Equation~\ref{eq:agg:distance}\;
            Compute the Weighting matrix $\mathbf{W}_{ij}$ as the inverse of the distance with Equation~\ref{eq:agg:inv_dist_weight}\;
        }
        Compute a normalisation factor $\mathbf{d}_i$ for the Weighting matrix using Equation~\ref{eq:agg:inv_dist_norm}\;
        Compute the new updated user embeddings without the contribution of the main client for the user with Equation~\ref{eq:agg:inv_dist_update}\;
        \eIf{i $\in$ $\mathcal{C}^{(r)}$}{
            Combine the new updated user embedding with the embedding of the anchor using a linear interpolation presented in Equation~\ref{eq:agg:inv_dist_aggegation}\;
        }{
            $\mathbf{e}^{(r)}_{u_i}$ = $\mathbf{e}^{\prime}_{u_i}$\;
        }
    }
    $\mathbf{E}_{u}^{(r)}$ = $\left\{\mathbf{e}^{(r)}_{u_0}, \mathbf{e}^{(r)}_{u_1}, \dots, \mathbf{e}^{(r)}_{u_\mathbf{n}}\right\}$\;

    \caption{Dist-FedAvg - Distance Based Averaging of User embeddings}
    \label{algo:agg:dist_fedavg}
\end{algorithm}

\subfour{Computation of Distance Matrix and Weight Matrix:} 
The first step in the aggregation involves computing the distance matrix between the different users, which we denote as $\mathbf{D}$. While in theory, it could be possible to use any similarity or distance measure to compute $\mathbf{D}$, we use the Minkowski distance, generalized as:
\begin{equation}\label{eq:agg:distance}
     \mathbf{D}_{ij} = \left(\sum_{k=0}^{K}|\mathbf{e}_{u_i,k} - \mathbf{e}_{u_j,k}|^{p}\right)^{\frac{1}{p}},
\end{equation}
where $K$ denotes the dimensionality of the user embedding $\mathbf{e}_{u_i}$. We calculate the Averaging Weight matrix $\mathbf{W}$ as the inverse of the distance using the following formula:
\begin{equation}\label{eq:agg:inv_dist_weight}
    \mathbf{W}_{ij} = 
    \begin{cases}
        \frac{1}{\mathbf{D}_{ij}},& \text{if } i \neq j \wedge u_j \in \mathcal{U}_i\\
        0, & \text{otherwise.}
    \end{cases}
\end{equation}
To ensure the weights lie between 0 and 1, we normalize them by a factor $\mathbf{d}_i$, defined as:\begin{equation}\label{eq:agg:inv_dist_norm}
    \mathbf{d}_i = \sum_{j=0}^{n} \mathbf{W}_{ij}.
\end{equation}
Then normalize our embeddings with the following equation: 
\begin{equation}\label{eq:agg:inv_dist_update}
    \mathbf{e}^{\prime}_{u_i} = \frac{1}{\mathbf{d}_i} . \sum_{j=0}^{n} \mathbf{e}_{u_{j,i}} \mathbf{W}_{ij}.
\end{equation}

\subfour{Handling the Anchor User and Interpolation:} 
In Equation~\ref{eq:agg:inv_dist_weight}, $\mathbf{W}_{ij} = 0$ when $i=j$, meaning the anchor user’s embedding is excluded from the computation of $\mathbf{e}^{\prime}_{i}$ in Equation~\ref{eq:agg:inv_dist_update}. To reintegrate the anchor user’s contribution, we perform a linear interpolation between $\mathbf{e}^{\prime}_{i}$ and the embeddings of the anchor user $\mathbf{e}_{u_{i,i}}$, with a parameter $\alpha$, as defined in the following equation:
\begin{equation}\label{eq:agg:inv_dist_aggegation}
    \mathbf{e}^{(r)}_{u_i} = \alpha \mathbf{e}_{u_{i,i}} + (1 - \alpha) \mathbf{e}^{\prime}_{u_i}.
\end{equation}

Note that a situation may arise where $u_i$ has no anchor client (in other words, $\mathbf{e}_{i,i}$ doesn't exist). This can happen in a given round $r$ if $u_i \not\in \mathcal{C}^{(r)}$. In other words, if $u_i$ wasn't selected for the local updates. In that case, we simply take $\mathbf{e}^{(r)}_{u_i} = \mathbf{e}^{\prime}_{u_i}$ and ignore Equation~\ref{eq:agg:inv_dist_aggegation} and default to $\alpha = 0$. This condition is expressed in line 9 of Algorithm~\ref{algo:agg:dist_fedavg}.

\subfour{Computation of $\alpha$ and Decay Strategies:}  
In Equation~\ref{eq:agg:inv_dist_aggegation}, we introduced a linear interpolation with a parameter $\alpha$; while the parameter could be provided as a hyperparameter, we propose an alternative way to compute it, one that solves the problem of weaknesses and lack of relevance in the distance in the early rounds. Since user embeddings are initialized randomly in early rounds, the distance between them is not representative of the similarity of users. Our solution is to implement an arithmetic decay strategy in which $\alpha$ starts at a high value and then decreases by a decay rate after a fixed number of rounds. Moreover, to avoid giving the expanded users higher weight than the anchor, to the eventual point where the anchor is completely ignored (when $\alpha$ is 0). We add a threshold value for $\alpha$, after which the decay will no longer be computed. The computation of $\alpha$ is presented in the following equation:
\begin{equation}\label{eq:agg:inv_dist_compute_alpha}
    \alpha = \operatorname{max}{\left(\alpha_{T}, \alpha_{0} -  \gamma \left \lfloor \frac{r}{z}\right \rfloor \right)}.
\end{equation}
Where $\alpha_{T}$ is the lower threshold for $\alpha$, $\alpha_{0}$ the initial value of $\alpha$, and $\gamma$ the decay. Finally, $r$ is the current round, and $z$ is the number of rounds to apply the decay. Another option to compute $\alpha$ is to implement a geometric decay such as: 
\begin{equation}\label{eq:agg:inv_dist_compute_geo_alpha}
    \alpha = \operatorname{max}{\left(\alpha_{T}, \alpha_{0}^{\gamma \left \lfloor \frac{r}{z}\right \rfloor} \right)}\\.
\end{equation}
Geometric decay decreases $\alpha$ more rapidly, which can more effectively leverage updates from other clients in early rounds. Alternatively, it is possible to simply add a certain number of warm-up rounds where $\alpha = 1$, thus $\mathbf{e}^{(r)}_{u_i} = \mathbf{e}_{u_{i,i}}$. Our experimental results show that while the decay can improve performance in some cases, such improvements are modest, and it is completely feasible to provide $\alpha$ directly as a hyperparameter.

%% file: sources/experiments.tex
\section{Experiments}\label{sec:experiments}

\subsection{Datasets and experimental setup}

\begin{table}[t]
    \caption{Datasets' statistics.}
    \centering
    \resizebox{\linewidth}{!}{
    \begin{tabular}{l@{\hskip 0.5in}r@{\hskip 0.2in}r@{\hskip 0.2in}r@{\hskip 0.2in}r}
        \hline
        \textbf{Dataset}       & \#Users & \#Items & \#Interactions & Sparsity \\
        \hline
        MovieLens-100k (ML-100k)~\cite{harperMovieLensDatasetsHistory2015}         & 943      & 1,682     & 100,000         & 93.69\%   \\
        MovieLens-1M (ML-1m)~\cite{harperMovieLensDatasetsHistory2015}           & 6,040    & 3,706     & 1,000,209       & 95.53\%  \\
        LastFM-2K~\cite{niJustifyingRecommendationsUsing2019}              & 1,892    & 17,632    & 92,834          & 99.72\%  \\
        AmazonMusic~\cite{houBridgingLanguageItems2024} & 5,541 & 3,568 & 64,706 & 99.67\%  \\
        FilmTrust~\cite{guoNovelEvidenceBasedBayesian2016} & 1,508 & 2,071	    & 35,497         & 98.86\%  \\
        \hline
    \end{tabular}}
    \label{tab:datasets}
\end{table}

For our experimental framework, we compare Dist-FedAvg with four existing state-of-the-art averaging strategies on five widely used datasets showcased in Table~\ref{tab:datasets}. The data preprocessing involves filtering out ratings below a designated threshold. Specifically, we use a threshold of 3 for the MovieLens datasets and Amazon Music, 2.5 for FilmTrust, and 10 interactions for LastFM, which doesn't have ratings but records interaction counts between users and artists. We further filter users with fewer than 15 interactions, and split the remaining ones into three sets, a training (70\%), validation (10\%), and testing (20\%) set using a random split~\cite{mengExploringDataSplitting2020}.

Our experimental setup is based on the simplified framework described in Section~\ref{sec:notation-problem-method} and Algorithm~\ref{algo:agg:central_server}\footnote{Code available at \url{https://anonymous.4open.science/r/Dist-FedAvg-CF3C/}}. For local training, we use LightGCN~\cite{heLightGCNSimplifyingPowering2020} with BPRLoss~\cite{rendleBPRBayesianPersonalized2009}. The local model is trained for a fixed number of local epochs in each FL round over a total of 100 rounds, employing an early stopping strategy with a patience of five rounds. The clients participating in the FL process ($\mathcal{C}^{(r)}$) are selected randomly at the beginning of each round.

\subsection{Comparison of Dist-FedAvg with Other Aggregation Methods. }

\begin{table*}[t]
    \caption{Comparison with other aggregation methods in terms of NDCG@10. 95\% confidence intervals are shown.}
    \label{tab:agg:evaluation-aggregation-metrics}
    \resizebox{\linewidth}{!}{
    \begin{tabular}{l|l|l|l|l|l}
        \hline
        \textbf{Avg. Meth.} & ML-100k & ML-1m & LastFM-2k & AmazonMusic & \multicolumn{1}{c}{FilmTrust} \\
        \hline
        FedAvg~\cite{mcmahanCommunicationEfficientLearningDeep2017}   & 0.1749$\pm$0.0113 & \textbf{0.1678$\pm$0.0047} & 0.1093$\pm$0.0071 & 0.0479$\pm$0.0078 & 0.3751$\pm$0.0176 \\
        SimpleAvg~\cite{perifanisFederatedNeuralCollaborative2022}      & 0.1766$\pm$0.0113 & 0.1659$\pm$0.0047 & 0.0987$\pm$0.0063 & 0.0575$\pm$0.0083 & 0.3402$\pm$0.0175 \\
        FedMedian~\cite{yinByzantineRobustDistributedLearning2018}       & 0.1720$\pm$0.0112 & 0.1456$\pm$0.0047 & 0.1100$\pm$0.0069 & 0.0567$\pm$0.0083 & 0.3223$\pm$0.0174  \\
        FedAtt~\cite{jiLearningPrivateNeural2019}                      & 0.1737$\pm$0.0114 & 0.1525$\pm$0.0047 & 0.0836$\pm$0.0062 & 0.0371$\pm$0.0068 & 0.3511$\pm$0.0179 \\
        \hline
        Dist-FedAvg                                                   & \textbf{0.1791$\pm$0.0117} & 0.1672$\pm$0.0047 & \textbf{0.1171$\pm$0.0077} & \textbf{0.0613$\pm$0.0081} &  \textbf{0.4027$\pm$0.018} \\ \hline
    \end{tabular}
    }
\end{table*}

\begin{table*}[t]
    \caption{Comparison with other aggregation methods in terms of HR@10. 95\% confidence intervals are shown.}
    \label{tab:agg:evaluation-aggregation-metrics_hr}
    \resizebox{\linewidth}{!}{
    \begin{tabular}{l|l|l|l|l|l}
        \hline
        \textbf{Avg. Meth.} & ML-100k & ML-1m & LastFM-2k & AmazonMusic & FilmTrust \\
        \hline
        FedAvg~\cite{mcmahanCommunicationEfficientLearningDeep2017}   & 0.6819$\pm$0.0298 & 0.6333$\pm$0.0122 & 0.4507$\pm$0.0227 & 0.2386$\pm$0.0296 & 0.8786$\pm$0.0242\\
        SimpleAvg~\cite{perifanisFederatedNeuralCollaborative2022}      & 0.6926$\pm$0.0295 & \textbf{0.6371$\pm$0.0121} & 0.3915$\pm$0.0223 & 0.2740$\pm$0.0311 & 0.8557$\pm$0.026 \\
        FedMedian~\cite{yinByzantineRobustDistributedLearning2018}       & 0.6777$\pm$0.0299 & 0.5472$\pm$0.0126 & \textbf{0.4804$\pm$0.0228} & 0.2689$\pm$0.0309 & 0.83$\pm$0.0278 \\
        FedAtt~\cite{jiLearningPrivateNeural2019}                      & 0.6734$\pm$0.0300 & 0.5815$\pm$0.0125 & 0.3986$\pm$0.0223 & 0.1932$\pm$0.0275 & 0.8457$\pm$0.0268 \\
        \hline
        Dist-FedAvg                                                   & \textbf{0.7053$\pm$0.0299} & 0.6345$\pm$0.0122 & 0.461$\pm$0.0225 & \textbf{0.2942$\pm$0.0317} & \textbf{0.89$\pm$0.0238} \\ \hline
    \end{tabular}
    }
\end{table*}

We compare Dist-FedAvg with four state-of-the-art aggregation methods: FedAvg~\cite{mcmahanCommunicationEfficientLearningDeep2017}, SimpleAvg~\cite{perifanisFederatedNeuralCollaborative2022}, FedMedian~\cite{yinByzantineRobustDistributedLearning2018}, and FedAtt~\cite{jiLearningPrivateNeural2019}. 
The results of the comparison are shown in Table~\ref{tab:agg:evaluation-aggregation-metrics} and~\ref{tab:agg:evaluation-aggregation-metrics_hr}. 
We performed basic hyperparameter tuning using Bayesian optimization for each method on every dataset presented in Table~\ref{tab:datasets}. The optimized hyperparameters include:
\begin{itemize}
    \item User aggregation hyperparameters: The hyperparameters of Dist-FedAvg and FedAtt, while other methods (FedAvg, SimpleAvg, FedMedian) require no specific hyperparameters. Specifically, for our method, we consider $p$ for the Minkowski distance in Equation~\ref{eq:agg:distance}, and the different decay parameters: the threshold $\alpha_{T}$, the decay value, the number of rounds to apply the decay, and a flag for whether to apply a geometric or arithmetic decay. We also experiment with giving $\alpha$ directly as a hyperparameter, as well as adding warmup rounds. Later experiments show that $\alpha$ can be set to a fixed value, allowing us to reduce the number of hyperparameters required for Dist-FedAvg.
    \item FL hyperparameters: These are the parameters required for the FL process in addition to the user aggregation, such as the number of selected clients, number of local iterations and item aggregation strategy.
    \item Local training hyperparameters: These represent the parameters needed for LightGCN, such as the learning rate, batch size, number of message passing layers, embedding size, etc.
\end{itemize}
Overall, Table~\ref{tab:agg:evaluation-aggregation-metrics} and~\ref{tab:agg:evaluation-aggregation-metrics_hr} shows that Dist-FedAvg outperforms the other aggregation methods, achieving higher NDCG@10 and HR@10 scores without significantly sacrificing training speed, execution time, privacy preservation, or simplicity.

\subsection{Ablation study and statistics on our method}

\begin{figure}[t]
    \centering
    \includegraphics[width=\columnwidth]{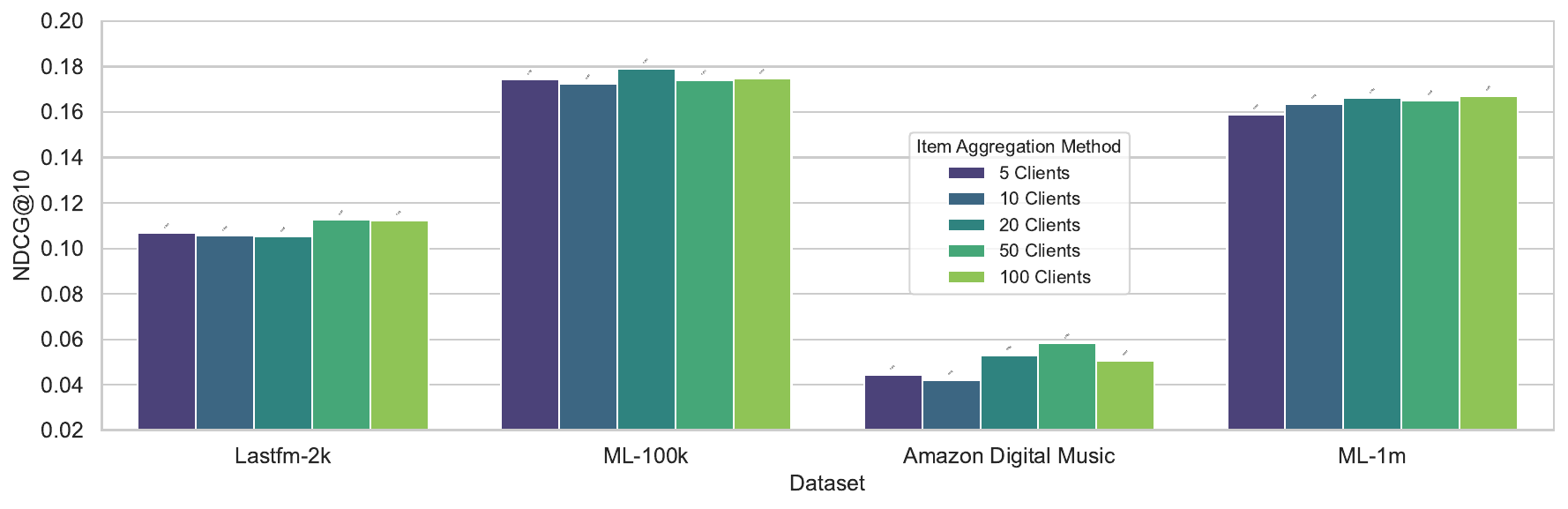}
    \caption{Difference in performance for Dist-FedAvg between different number of selected clients.}
    \label{fig:comparison_selected_clients}
\end{figure}

\subsubsection{Impact of number of selected clients in a federated round:} 
We evaluate Dist-FedAvg using varying numbers of selected clients, as illustrated in Figure~\ref{fig:comparison_selected_clients}. As expected, performance tends to improve with an increasing number of clients, but the benefits plateau around 20 or 50 clients. We can also note that the datasets with higher number of users requires a slightly higher number of clients. This observation is particularly important because Dist-FedAvg's efficiency depends on updating user embeddings across multiple clients. However, training with a very large number of clients is not always possible and is often counterproductive in FL. As such, finding the right balance for the number of clients for Dist-FedAvg is important.

\subsubsection{Impact of alpha Decay and thresholds to Dist-FedAvg:}

\begin{figure}[t]
  \centering
  \begin{minipage}{0.48\textwidth}
    \centering
    \includegraphics[width=\linewidth]{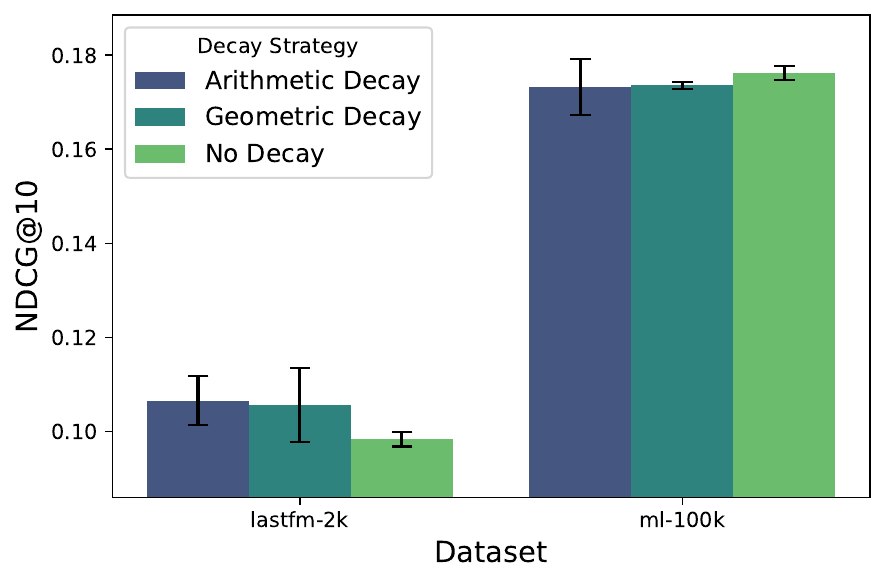}
    \caption{Comparison of geometric, arithmetic, and no decay strategies for computing $\alpha$.}
    \label{fig:comparison_decay}
  \end{minipage}
  \hfill
  \begin{minipage}{0.48\textwidth}
    \centering
    \includegraphics[width=\linewidth]{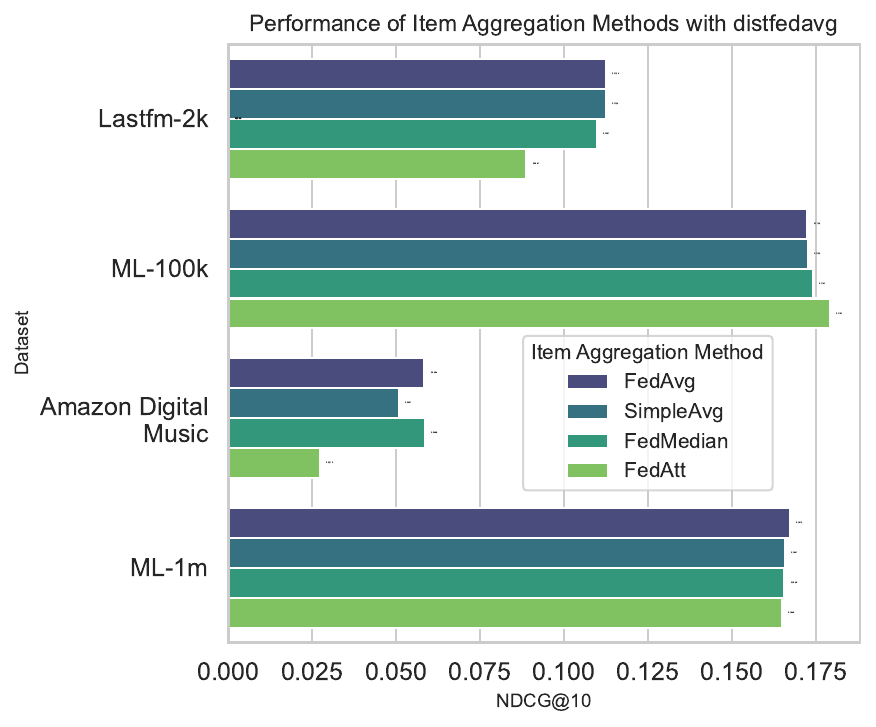}
    \caption{Comparison of item Aggregation methods paired with Dist-FedAvg.}
    \label{fig:comparison_item_agg_distfedavg}
  \end{minipage}
\end{figure}

A key component of Dist-FedAvg is the decay of the interpolation parameter $\alpha$.
As explained in Section~\ref{sec:proposed-method}, in early rounds, the user embeddings  are less representative and do not effectively capture user similarity. To address this, we introduced a decay for $\alpha$, where it starts at 1, and is gradually reduced to a predefined threshold over training.
We propose two decay strategies for $\alpha$: an arithmetic decay (Equation~\ref{eq:agg:inv_dist_compute_alpha}) and a geometric decay (Equation~\ref{eq:agg:inv_dist_compute_geo_alpha}). In Figure~\ref{fig:comparison_decay}, we evaluate whether a decay is truly needed or whether $\alpha$ can be provided as a hyperparameter reducing Dist-FedAvg's complexity. Furthermore, we evaluate which decay type is better suited for Dist-FedAvg. The results show that the decay doesn't always improve the results. Surprisingly, ML-100k achieves better results without decay, despite the problem of lack of relevance of user embeddings in early rounds. However, we must note that the user embeddings have been trained for 20 local epochs in each round, and we might not achieve good results with a lower number of local epochs per round without a decay, but this can in theory be circumvented by adding warmup rounds where we simply ignore the linear interpolation in Equation~\ref{eq:agg:inv_dist_aggegation}. We can also see that overall for both LastFM and Movie Lens, the geometric decay provides similar results to the arithmetic decay.

\subsubsection{Aggregation of item embeddings:}

\begin{figure}[t]
  \centering
  \includegraphics[width=\linewidth]{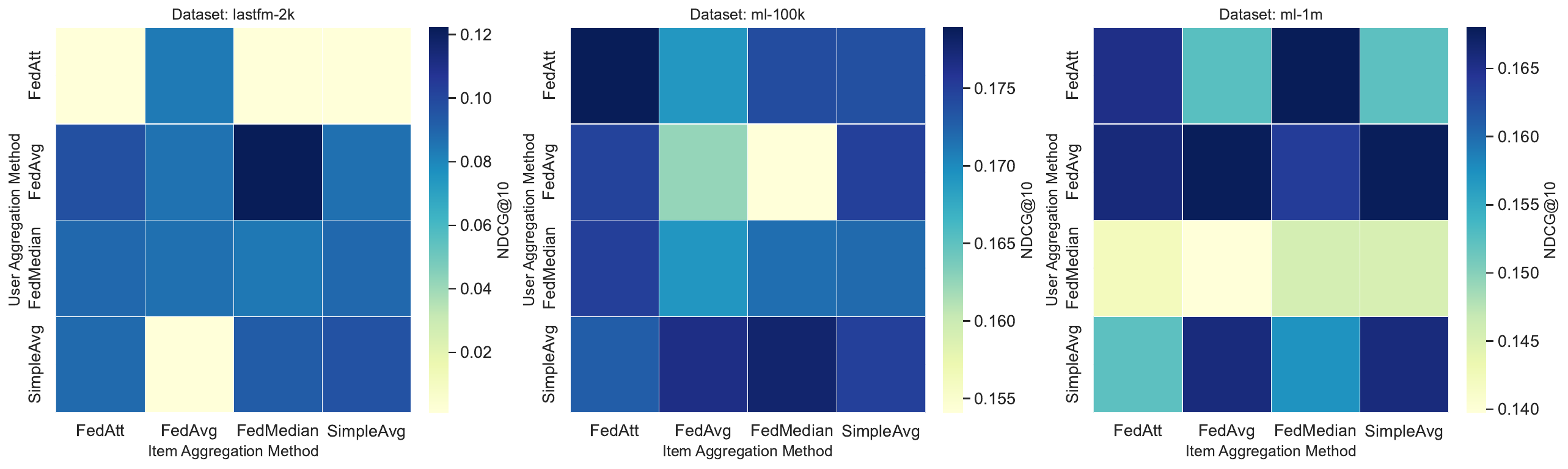}
  \caption{Heatmap analyzing the impact of using different user (y-axis) and item (x-axis) aggregation methods. The heatmap intensity represents the NDCG@10 score when using a combination of user and item aggregation methods }
  \label{fig:relations_aggregation_methods}
\end{figure}

While Dist-FedAvg effectively aggregates user embeddings, it cannot be directly applied to item embeddings. This raises concerns that employing different aggregation methods for user and item embeddings might negatively affect performance. We investigate this concern in Figure~\ref{fig:relations_aggregation_methods}, where we visualize in a heatmap, the relation between user and item embedding aggregation methods, our goal is to determine whether using the same method for both user and item aggregation leads to better results. Figure~\ref{fig:relations_aggregation_methods} clearly shows that using the same aggregation method for both user and item embeddings does not necessarily correlate with higher performance; in fact, employing different methods for each often yields better results. While this would mean that Dist-FedAvg can be used as is for aggregating user parameters, we argue that there is still merit in adapting Dist-FedAvg for item parameters, which can be the subject of future work.
For the time being, we aim to experimentally determine which aggregation method is best paired with Dist-FedAvg. Figure~\ref{fig:comparison_item_agg_distfedavg} shows the performance of each item aggregation method when used alongside Dist-FedAvg. While the difference is often small, we can see that simpler methods such as FedAvg and SimpleAvg seem to give better results overall.

%% file: sources/conclusion.tex
\section{Conclusions and Future Work}\label{sec:conclusion}
In this paper, we propose Dist-FedAvg, a novel aggregation method for graph-based federated recommendation systems. Dist-FedAvg computes aggregation weights using distances between expanded users in local user-item interaction graphs. Then combine these weighted embeddings with the anchor user's (i.e., the central user within a client’s local graph) parameters via a linear interpolation. Our method is intended to be used as an add-on with minimal changes to existing graph federated recommendation frameworks that rely on user embedding aggregation.
Our experiments for Dist-FedAvg on five RS datasets show great promise for the method. Our method performs better than all tested aggregation methods, with no significant sacrifice on training speed or privacy preservation.
Future work could integrate and evaluate Dist-FedAvg within existing graph frameworks (e.g., FedPerGNN) or extend it to non-graph-based federated recommendation frameworks. Additionally, exploring the aggregation of item embeddings based on user-derived weights represents a promising direction for further research. 